\documentclass[lettersize,journal]{IEEEtran}
\usepackage{amsmath,amsfonts}
\usepackage{algorithmic}
\usepackage{algorithm}
\usepackage{array}
\usepackage[caption=false,font=normalsize,labelfont=sf,textfont=sf]{subfig}
\usepackage{textcomp}
\usepackage{stfloats}
\usepackage{url}
\usepackage{verbatim}
\usepackage{graphicx}
\usepackage{cite}
\usepackage[colorlinks,
            linkcolor=red,
            anchorcolor=blue,
            citecolor=green]{hyperref}
\usepackage{bm}

\usepackage{amssymb}
\usepackage{multirow}
\usepackage[normalem]{ulem}
\usepackage{booktabs}
\usepackage{colortbl}
\usepackage{color}
\usepackage{xcolor}
\usepackage{tabu}

\begin{document}

\title{Unsupervised Domain Adaptation via Domain-Adaptive Diffusion}

\author{Duo Peng, Qiuhong Ke, Yinjie Lei and Jun Liu
\thanks{Corresponding author: Jun Liu.}
\thanks{This research/project is supported by the National Research Foundation, Singapore under its AI Singapore Programme (AISG Award No: AISG2-PhD-2022-01-027[T]).}
\thanks{Duo Peng and Jun Liu are with the Information Systems Technology and Design Pillar,
Singapore University of Technology and Design, Singapore 487372 (e-mail:
duo\_peng@mymail.sutd.edu.sg, jun\_liu@sutd.edu.sg).}
\thanks{Qiuhong Ke is with the Department of Data Science and AI, Monash University, Parkville, VIC 3010, Australia (email: qiuhong.ke@monash.edu).}
\thanks{Yinjie Lei is with the College of Electronics and Information Engineering, Sichuan University, Chengdu 610065, China (email: yinjie@scu.edu.cn).}
}

\markboth{IEEE TRANSACTIONS ON IMAGE PROCESSING}%
{Shell \MakeLowercase{\textit{et al.}}: A Sample Article Using IEEEtran.cls for IEEE Journals}

\maketitle

\begin{abstract}
Unsupervised Domain Adaptation (UDA) is quite challenging due to the large distribution discrepancy between the source domain and the target domain. Inspired by diffusion models which have strong capability to gradually convert data distributions across a large gap, we consider to explore the diffusion technique to handle the challenging UDA task. However, using diffusion models to convert data distribution across different domains is a non-trivial problem as the standard diffusion models generally perform conversion from the Gaussian distribution instead of from a specific domain distribution. Besides, during the conversion, the semantics of the source-domain data needs to be preserved for classification in the target domain. To tackle these problems, we propose a novel Domain-Adaptive Diffusion (DAD) module accompanied by a Mutual Learning Strategy (MLS), which can gradually convert data distribution from the source domain to the target domain while enabling the classification model to learn along the domain transition process.   Consequently, our method successfully eases the challenge of UDA by decomposing the large domain gap into small ones and gradually enhancing the capacity of classification model to finally adapt to the target domain. Our method outperforms the current state-of-the-arts by a large margin on three widely used UDA datasets.
\end{abstract}

\begin{IEEEkeywords}
Unsupervised domain adaptation, Diffusion model, Transfer learning, Image classification.
\end{IEEEkeywords}

\section{Introduction}
\IEEEPARstart{D}{eep}  Neural Networks (DNNs) \cite{he2016deep,krizhevsky2012imagenet,simonyan2014very} have achieved great success in various types of visual recognition tasks. Despite the success already achieved, the performance of DNNs highly relies on large amounts of annotated training data which is often prohibitively laborious and time-consuming to collect \cite{deng2009imagenet,everingham2015pascal}. One alternative that could mitigate this constraint is to leverage the off-the-shelf labeled data (namely source-domain data). However, the model trained with source-domain data often experiences obvious performance drop when applied to a target domain which has discrepant distributions \cite{li2020model,pan2019transferrable,pinheiro2018unsupervised}. To deal with this issue, Unsupervised Domain Adaptation (UDA) \cite{wilson2020survey} is presented which transfers knowledge from a labeled source domain to a different unlabeled target domain. UDA is a very challenging problem as there can be a large discrepancy between the two domains, resulting in great difficulties in knowledge transfer. 

Recently, Denoising Diffusion Probabilistic Models (DDPMs) \cite{ho2020denoising,sohl2015deep} (represented as \textit{diffusion models} for brevity) have achieved outstanding performance, showing a trend to surpass current state-of-the-art Generative Adversarial Networks (GANs) \cite{NIPS2014_5ca3e9b1} in various research fields, such as image editing \cite{meng2021sdedit}, likelihood estimation \cite{kingma2021variational}, and realistic-looking image generation \cite{dhariwal2021diffusion,nichol2021improved}. Generally, diffusion models can successfully convert data from the Gaussian noise into natural images, even though there are large distribution discrepancies between these two types of data. The great success of diffusion models can be attributed to its ability to decompose the large-gap conversion into multiple small-gap ones and thereby complete them gradually, making the distribution conversion smoother and easier. 

Inspired by the powerful distribution conversion capability of DDPM, we consider to leverage diffusion models to tackle the challenging UDA task, in which there is also a large gap between data distributions (of the source and target domains). Based on this motivation, we propose to leverage the diffusion technique to achieve a cross-domain distribution conversion, where the source-domain distribution can be gradually and smoothly converted into the target domain. However, this is a non-trivial problem owing to the following challenges: (1) Traditional diffusion models mainly focus on converting data from a simple distribution (e.g., Gaussian) to a target data distribution, while UDA requires the model to start with a specific source-domain distribution.  (2) During source-to-target conversion, the semantic (category) information of the source-domain data needs to be retained, in order to enable domain-adaptive training of the classification model for classifying the target-domain data. In this paper, we propose a Domain-Adaptive Diffusion (DAD) module accompanied with a Mutual Learning Strategy (MLS) to address these challenges.

Specifically, inspired by the distribution conversion mechanism of diffusion models, we carefully design a DAD module to slightly and iteratively change the source-domain distribution towards the target domain. 
As shown in Figure \ref{fig_3} (b), in our DAD module, the source-domain distribution is progressively diffused over multiple steps, producing a series of distributions under different diffusion levels (see yellow blocks in Figure \ref{fig_3} (b)). Then the DAD module converts (concentrates) each diffused distribution with the same number of steps for eliminating the diffusion (blue blocks). As the DAD module is trained to concentrate distributions towards the target domain, by concentrating the distribution over different levels (steps), our framework can simulate the cross-domain distribution transition process. As shown in the bottom part of Figure \ref{fig_3} (b), the DAD module provides a sequence of transitional features whose distributions are changing gradually. In the sequence, every two neighboring transitional features show a minor distribution discrepancy, decomposing the large domain gap into small ones and thus enabling a much easier domain adaptation.

During the source-to-target simulation process, we train the classification model on each simulated transitional distribution, in order to gradually enlarge the distribution scope learned by the classification model. With this manner, the model can finally adapt to the target domain. Note that the classification model’s learning on each transitional feature distribution relies on the same labels of the source domain. This requires the semantics of the feature maps to be preserved during distribution transition.
To this end, we further propose a Mutual Learning Strategy (MLS) where the \textbf{DAD module} and the \textbf{classification model} are alternately learned from each other, aiming to gradually boost the classification model's adaption capacity with the preservation of semantics of features.
Our proposed MLS involves two learning directions (namely, “Classification-to-DAD” and “DAD-to-Classification”). As shown in the bottom part of Figure \ref{fig_3} (a), the “Classification-to-DAD” learning (red dashed arrow) leverages the classification model, which is already trained on former distributions, to constrain feature semantics of the next distribution via updating the DAD module. In turn, the “DAD-to-Classification” learning (red solid arrow) utilizes the updated DAD module to promote the classification model to learn a new distribution, aiming to enhance its adaption capacity. We iteratively alternate the above two learning directions, driving the classification model to generalize across small gaps, finally covering the target domain.

It is worth mentioning that both DAD and MLS are only applied at the training phase for training the classification model. This means DAD and MLS are removed during testing. Therefore, our method brings no additional computation expense for inference. Extensive experiments demonstrate that our method achieves state-of-the-art performance on three widely used UDA benchmarks, such as Office-31 \cite{saenko2010adapting}, Office-Home \cite{venkateswara2017deep}, and VisDA-2017 \cite{peng2018visda}, and outperforms other methods by a large margin. 

We summarize our contributions as follows:
\begin{itemize}
\item 
We design a novel Domain-Adaptive Diffusion (DAD) module taking advantage of the strong distribution conversion capability of DDPM, 
which bridges the source and target domains via a sequence of simulated distributions with small distribution discrepancies in adjacent distributions, thus alleviating the challenge of UDA.
\vspace{3mm}

\item 
We devise a Mutual Learning Strategy (MLS) to enhance the domain-adaptive capacity of the classification model during the distribution transition process.
\vspace{3mm}

\item 
Our method achieves state-of-the-art performance on various UDA benchmarks.
\end{itemize}

\section{Related Work}

\subsection{Unsupervised Domain Adaptation (UDA).} 
Collecting a plenty of annotated training data for each domain of interest is often  costly and time-consuming. To address this issue, various UDA methods have been proposed.
Saenko et al. \cite{saenko2010adapting} firstly gave an exploration on UDA, obtaining much attention and fruitful results \cite{wang2012multi,yang2007cross}. Techniques developed based on deep networks can be mainly categorized into three types: feature-level adaptation, image-level adaptation, and both feature and image levels adaptation.
Feature-level adaptation aims to align features by either directly minimizing the feature distance \cite{long2015learning,sun2016deep,long2017deep} or implicitly narrowing down the gap between source and target distributions (e.g., via adversarial learning \cite{kang2018deep,long2018conditional,zhang2019domain,ganin2016domain,tzeng2017adversarial,russo2018source}).
Meanwhile, 
image-level adaptation attempts to remove differences by either translating source images to target images \cite{murez2018image,ramirez2018exploiting} or utilizing mixup techniques \cite{zhang2017mixup} to mix images from the two domains \cite{verma2019manifold,berthelot2019mixmatch,wu2020dual}.
There are also methods taking both the feature and image levels into consideration \cite{hoffman2018cycada,wu2018dcan}. 

Although the methods mentioned above have achieved remarkable progress, they still face performance limitations due to the large domain gap. Inspired by the diffusion models that are capable of handling large-gap distribution conversion and the particular challenges in UDA, from a novel perspective, we propose a framework with a new Domain-Adaptive Diffusion (DAD) module and a Mutual Learning Strategy (MLS) to deal with the challenging UDA problem.
To the best of our knowledge, this is the first UDA framework with a novel diffusion model design, which leverages the newly designed DAD and MLS and achieves superior performance on multiple benchmarks.

\subsection{Denoising Diffusion Probabilistic Models (DDPMs).} 
DDPMs \cite{ho2020denoising,sohl2015deep}, also known as diffusion models for brevity, are a class of generative models inspired by non-equilibrium thermodynamics. Diffusion models generally learn a parameterized Markov chain to gradually denoise from an original common distribution to a specific data distribution. It is first proposed by Sohl-Dickstein et al. \cite{ho2020denoising}, and has attracted much attention recently due to its state-of-the-art performance on various tasks, including image generation \cite{zhou20213d,dhariwal2021diffusion}, image editing \cite{meng2021sdedit}, point cloud completion \cite{Choi2021Ilvr} and stochastic trajectory prediction \cite{gu2022stochastic}. Based on the basic theory of DDPM, there have been research works aiming to improve diffusion models in different aspects. Song et al. \cite{song2020score} proposed to leverage techniques from stochastic differential equations to improve the sample quality. Song et al. \cite{song2020denoising} and Nichol et al. \cite{nichol2021improved} proposed methods to improve sampling (inference) speed. Nichol et al. \cite{nichol2021improved} and Saharia et al. \cite{saharia2022image} successfully introduced upsampling into diffusion models, obtaining promising results.

Different from the above works, here we propose a new diffusion-based UDA framework that converts data from a specific source-domain distribution to the target domain, simulating the process of domain transition for handling the challenging UDA task. A novel learning strategy is also proposed to support the training of our UDA framework.

\begin{figure*}[t]
\begin{centering}
\includegraphics[scale=0.145]{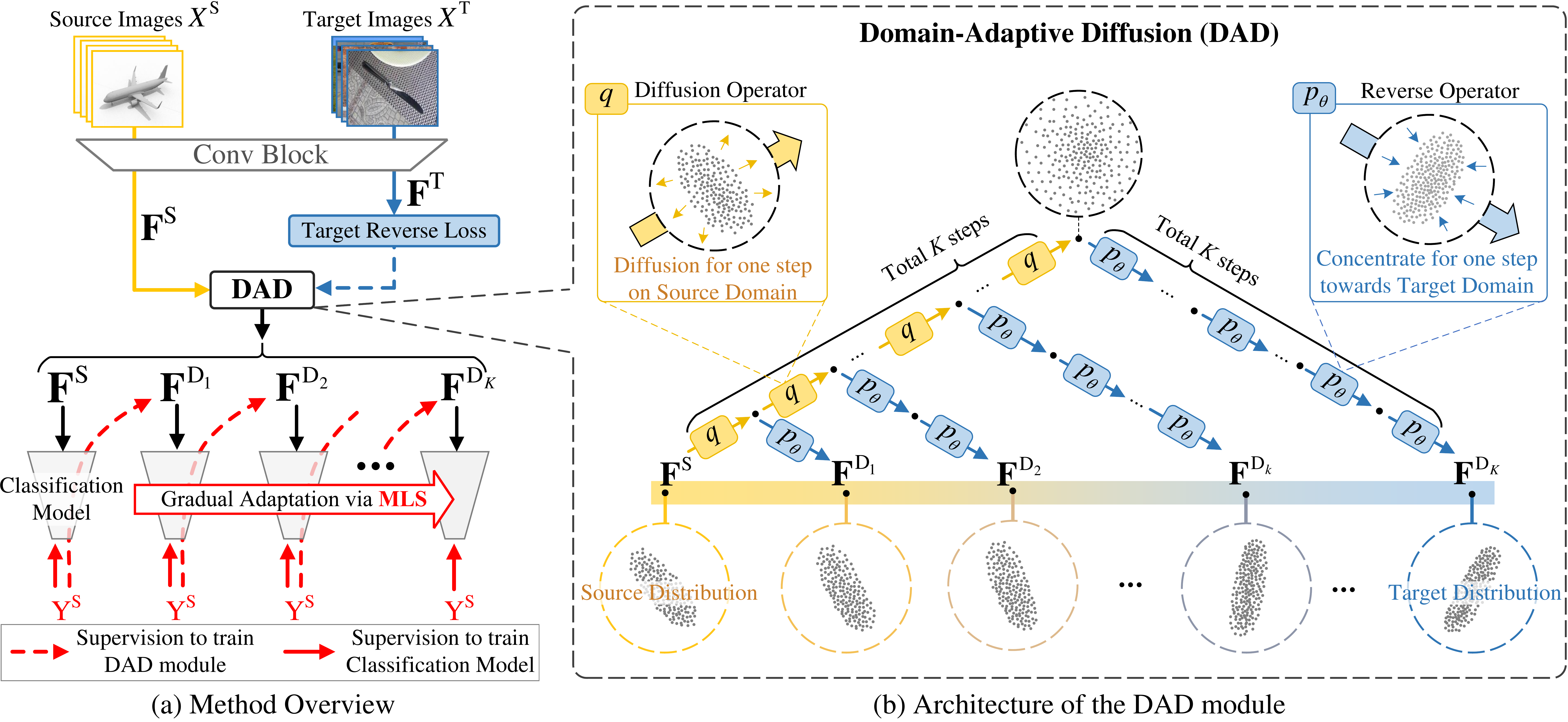} 
\par\end{centering}
\vspace{-2mm}
\caption{Illustration of our framework. \textbf{(a)} In our method, the Conv Block maps the source and target images into features $\textbf{F}^{\rm S}$ and $\textbf{F}^{\rm T}$, respectively. We feed $\textbf{F}^{\rm S}$ into the DAD module to simulate the cross-domain distribution transition process. Then, we use $\textbf{F}^{\rm T}$ with the Target Reverse Loss to empower the DAD module to convert distributions towards the target domain. For a better cross-domain distribution conversion, we further propose a Mutual Learning Strategy (MLS) to preserve semantics of each transitional distribution. In MLS, the classification model and the DAD module are learned from each other. Specifically, the classification model which is well-trained on former distributions is used to constrain the feature semantics of the next (new) distribution via updating the DAD module (red dashed arrow). After updating the DAD module, the semantic-preserved features of the next distribution can guide the classification model's learning on this new distribution (red solid arrow). MLS alternates the above two training schemes, gradually improving the classifier's adaptive capacity towards the target domain with preservation of feature semantics. 
\textbf{(b)} The DAD module contains the diffusion operator $\bm{q}$ and the reverse operator $\bm{p}_\theta$. Given the source features $\textbf{F}^{\rm S}$, DAD can gradually change its distribution into the target domain by iteratively diffusing on source domain and concentrating towards target domain.
}
\label{fig_3} 
\vspace{-5mm}
\end{figure*}

\section{Revisiting Diffusion Models}\label{sec:revisiting}

Our method is proposed based on diffusion models (i.e., DDPM). Below, we briefly revisit DDPM. A standard DDPM generally contains two basic units: the diffusion operator $\bm{q}$ and the reverse operator $\bm{p}_\theta$. Specifically, the diffusion operator $\bm{q}$ is used to diffuse the data distribution by adding perturbations, while the reverse operator $\bm{p}_\theta$ is utilized to concentrate the data distribution by eliminating the perturbations. In DDPM, applying the diffusion operator $\bm{q}$ (or reverse operator $\bm{p}_\theta$) once can lead to diffusion (or concentration) for one step, which changes the distribution slightly. Given the feature map $\textbf{F}$ under the distribution $\mathcal{A}$, the diffusion operator $\bm{q}$ is iteratively used to gradually diffuse the data $\textbf{F}$ for $K$ steps, obtaining the diffusion process, i.e., $\{\textbf{F} , \textbf{F}_1 , \textbf{F}_2 ,..., \textbf{F}_K\}$. Mathematically, the diffusion operator $\bm{q}$ can be defined as:
    \begin{equation}
    \label{q}
    q(\textbf{F}_k|\textbf{F}_{k-1})=\mathcal{N}(\textbf{F}_k;\sqrt{1-\beta_k}\textbf{F}_{k-1},\beta_k\textbf{I}),
    \end{equation}
where $k\in \{1,2,...,K\}$. 
$\textbf{F}_{k-1}$ is the feature map already diffused for $(k-1)$ steps which is the input of $\bm{q}$, and $\textbf{F}_k$ is the output of the diffusion operator $\bm{q}$. $\{\beta_{1}, \beta_{2}, \cdots, \beta_{K}\}$ are fixed variance schedulers that control the scale of the injected perturbation. When $K$ is large enough, $\textbf{F}_K$ can be approximately regarded as the Gaussian-distributed data. 

Different from diffusion operator $\bm{q}$ that is defined with fixed parameters, the reverse operator $\bm{p}_\theta$ is defined with learnable parameters $\theta$. The goal of the standard DDPM is to train the reverse operator $\bm{p}_{\theta}$ to gradually convert the Gaussian noise $\textbf{F}_K$ into data $\textbf{F}$ of desired distribution, forming the reverse process $\{\textbf{F}_K , \textbf{F}_{K-1} , \textbf{F}_{K-2} ,..., \textbf{F}\}$. The reverse operator $\bm{p}_\theta$ is formulated as:
    \vspace{-0.0 mm}
    \begin{equation}
    \label{p_theta}
     p_\theta(\textbf{F}_{k-1}|\textbf{F}_{k})=\mathcal{N}(\textbf{F}_{k-1};\mu_\theta(\textbf{F}_k,k ),\beta_k\textbf{I}),
    \vspace{-0.0 mm}
    \end{equation}
where $\textbf{F}_{k}$ is the input feature map of the reverse operator $\bm{p}_\theta$, and $\textbf{F}_{k-1}$ is the output. $\mu_\theta(\textbf{F}_k,k)$ is the mean value of the output feature distribution, which can be predicted by the reverse operator $\bm{p}_{\theta}$ as:
    \vspace{-0 mm}
    \begin{equation}
    \label{mu_theta}
    \mu_\theta(\textbf{F}_k,k )=\frac{1}{\sqrt{\alpha _k} }\big{(}\textbf{F}_{k}-\frac{\beta_k}{\sqrt{1-\bar{\alpha}_k}}f_\theta(\textbf{F}_k,k) \big{)},
    \vspace{-0.0 mm}
    \end{equation}
where $f_\theta$ is the learnable neural network within the reverse operator $\bm{p}_\theta$, $\alpha_k=1-\beta_k$, and $\overline { \alpha} _ { k } =\prod _ { s = 1 } ^ { k }\alpha_s$. In each reverse (concentration) step, the reverse operator $\bm{p}_\theta$ uses its neural network $f_\theta$ to first predict the mean value $\mu_\theta(\textbf{F}_k,k)$ of the output feature map, and then uses the predicted mean value to formulate the output distribution via Eqn. \ref{p_theta}. Note that the network $f_{\theta}$ is shared over all reverse steps.

In order to train the reverse operator $\bm{p}_\theta$ to convert any random Gaussian noise into features with distribution $\mathcal{A}$, the features $\textbf{F}$ (under distribution $\mathcal{A}$) are first diffused by the diffusion operator $\bm{q}$ for $K$ steps, obtaining its diffusion process. Then all data in the diffusion process are used as supervision to guide the learning of the reverse operator. Based on the derivation in \cite{ho2020denoising}, the reverse learning loss is defined as:
    \vspace{-0 mm}
    \begin{equation}
    \label{L_DAD}
    \mathcal{L}_{\rm RL}(\textbf{F})=\mathbb{E}_q \big{[}-{\rm log}\ p(\textbf{F}_K)-\sum_{k=1}^{K}{\rm log}\ \frac{\bm{p}_\theta(\textbf{F}_{k-1}|\textbf{F}_{k})}{\bm{q}(\textbf{F}_{k}|\textbf{F}_{k-1})} \big{]},
    \vspace{-0.0 mm}
    \end{equation}
where $p(\textbf{F}_K)$ is the initial Gaussian distribution. After the reverse learning, given a random Gaussian noisy feature map, the reverse operator $\bm{p}_{\theta}$ can gradually convert the feature map into the desired distribution $\mathcal{A}$.

\section{Proposed Approach}

We address the problem of UDA, where we have access to the source data $X^{\rm S}$ with labels $Y^{\rm{S}}$, and the target data $X^{\rm T}$ without labels. Our goal is to use the above data to train a classification network that works well in the target domain. In light of diffusion models' powerful capability in converting distribution across a large gap via multiple small gaps, in this paper, we propose to leverage the diffusion design to handle cross-domain distribution conversion, aiming to better tackle the challenging adaptation problem caused by the large domain gap. To this end, we propose a novel UDA framework with a Domain-Adaptive Diffusion (DAD) module and a Mutual Learning Strategy (MLS).

The overview of our framework is shown in Figure \ref{fig_3} (a). In this paper, we insert our DAD module after the first convolution stage of the backbone network (e.g., ResNet-50), separating the backbone network into two parts. For brevity, the shallow part before the DAD module is named ``Conv Block'' which maps the source images $X^{\rm S}$ and target images $X^{\rm T}$ into features $\textbf{F}^{\rm S}$ and $\textbf{F}^{\rm T}$, respectively. The deep part after the DAD module is called ``classification model'' that maps DAD-outputted features into classification results.

Specifically, we conduct UDA with the following scheme. \textbf{(1)} Given extracted source features $\textbf{F}^{\rm S}$ (under source-domain distribution $\mathcal{S}$) and target features $\textbf{F}^{\rm T}$ (under target-domain distribution $\mathcal{T}$), our DAD module learns to simulate the cross-domain distribution transition process, obtaining a sequence of transitional features $\{\textbf{F}^{\rm S}, \textbf{F}^{{\rm D}_1}, \textbf{F}^{{\rm D}_2},..., \textbf{F}^{{\rm D}_K}\}$ with gradual-changing distributions, where $\textbf{F}^{{\rm D}_K}$ follows the target-domain distribution $\mathcal{T}$. \textbf{(2)} During the simulation process using the DAD module, we enable the classification model and the DAD module to learn from each other, which follows the Mutual Learning Strategy (MLS). As shown in Figure \ref{fig_3} (a), at iteration $k$, the classification model that is already trained on a series of former features $\{\textbf{F}^{\rm S}, \textbf{F}^{{\rm D}_1},..., \textbf{F}^{{\rm D}_{k-1}}\}$ ($k\in \{1,2,...,K\}$), is utilized to supervise the feature semantics of the next features $\textbf{F}^{{\rm D}_{k}}$ via updating the DAD module (red dashed arrow). In turn, the updated DAD module provides semantic-preserved features $\textbf{F}^{{\rm D}_{k}}$ to further help the classification model to learn on a new distribution (red solid arrow). Thus, after the alternate learning iteration $k$, the classification model can perform well on features $\{\textbf{F}^{\rm S}, \textbf{F}^{{\rm D}_1},..., \textbf{F}^{{\rm D}_{k-1}}, \textbf{F}^{{\rm D}_{k}}\}$, which is closer to the target domain. During the whole alternate learning iterations with MLS (i.e., increasing $k$ from 1 to $K$), the model's classification adaptive capacity is gradually enhanced and enlarged, and finally, our model can yield good performance in the target domain.
Next, we detail the architecture of our DAD module and describe the training objective including MLS.

\subsection{Domain-Adaptive Diffusion (DAD)}\label{sec:DAD}

Inspired by DDPM that handles distribution conversion by decomposing large-gap conversion into multiple small-gap ones, our DAD module is designed to utilize diffusion technique to model the source-to-target distribution transition process, aiming to decompose the large domain gap into small ones.
Specifically, our DAD module contains two basic units: the diffusion operator $\bm{q}$ and the reverse operator $\bm{p}_\theta$ which have the same definition as diffusion models.
Different from standard diffusion models separating the diffusion process and reverse process apart, we utilize both the diffusion and reverse operators at each iteration step during the transition process to bridge distributions of two diferent domains for UDA.
Note that a single step of the diffusion (or reverse) operator only diffuses (or concentrates) the distribution slightly.

As shown in Figure \ref{fig_3}(b), we utilize the diffusion operator $\bm{q}$ to diffuse source features $\textbf{F}^S$ for a total of $K$ steps (see the yellow blocks in Figure \ref{fig_3}(b)), obtaining the features under different levels of diffusion. For each diffused source feature map in the diffusion process, we then apply the reverse operator $\bm{p}_\theta$ with the corresponding number of steps to convert (concentrate) the diffused distribution into a specific distribution (blue blocks in Figure \ref{fig_3}(b)). Note that we design a training strategy (detailed in Sec. \ref{sec:MLS}) to train the reverse operator $\bm{p}_\theta$ in our DAD module, giving it the ability to convert (concentrate) distribution towards the target domain rather than backing to the source domain. Therefore, applying the single reverse operator $\bm{p}_\theta$ once can slightly concentrate the diffused distribution, encouraging the concentrated feature distribution to progressively approach the target domain.

More specifically, for source features which have been diffused for $k$ $(k\in \{1,2,..., K\})$ steps, we use the reverse operator $\bm{p}_\theta$ to concentrate the diffused features with the same number of steps (i.e., $k$), obtaining the domain-transitional features $\textbf{F}^{{\rm D}_k}$, as shown in Figure \ref{fig_3} (b). Thus, when $k$ is small, the DAD module makes minor changes to the feature distribution, obtaining the domain-transitional features with small differences compared to the source-domain distribution (e.g., $\textbf{F}^{{\rm D}_1}$ and $\textbf{F}^{{\rm D}_2}$ in Figure \ref{fig_3} (b)). 
However, with $k$ growing from 1 to $K$, the source features are increasingly diffused, and meanwhile, more and more target reverse operations are carried out, making the output features gradually approach the target domain and thus simulating the domain transition process $\{\textbf{F}^{\rm S}, \textbf{F}^{{\rm D}_1},\textbf{F}^{{\rm D}_2},..., \textbf{F}^{{\rm D}_K}\}$.

Given the source features $\textbf{F}^S$, the simulation of the feature map $\textbf{F}^{{\rm D}_k}$ $(k\in \{1,2,...K\})$ with the DAD module can be formulated as:
    \vspace{-0.0 mm}
    \begin{equation}
    \label{DAD}
    \textbf{F}^{{\rm D}_k}=f^k_{\rm D}(\textbf{F}^{\rm S})={\rm Rev}\big{(} {\rm Dif}(\textbf{F}^{\rm S} , k)\ ,k \big{)},
    \vspace{-0 mm}
    \end{equation}
where $f^k_{\rm D}$ denotes the DAD module that handles the simulation of features $\textbf{F}^{{\rm D}_k}$, ${\rm Dif}(\cdot,k)$ denotes using the diffusion operator $\bm{q}$ to diffuse for $k$ consecutive steps, and ${\rm Rev}(\cdot,k)$ represents using the reverse operator $\bm{p}_\theta$ to reverse for $k$ consecutive steps.

\subsection{Training Objective}\label{sec:MLS}

After formulating our DAD module, below we describe the training losses.

\textbf{Source training loss.} Before adaptation, we first train the backbone network on the source domain images $X^{\rm S}$ with labels $Y^{\rm S}$, using the source training loss:
    \vspace{-0 mm}
    \begin{equation}
    \label{L_ST}
    \mathcal{L}_{\rm ST}=\mathcal{L}_{cls}\Big{(} f_{\rm C} \big{(} f_{\rm E} (X^{\rm S})\big{)},Y_{\rm S}\Big{)},
    \vspace{-0 mm}
    \end{equation}
where $f_{\rm E}$ and $f_{\rm C}$ denote the encoding Conv Block 
and the classification model, respectively. $\mathcal{L}_{cls}$ represents the cross-entropy loss that is widely used in classification tasks.

\textbf{Target reverse loss.} After training the backbone network on the source domain, we then train the DAD module to give DAD an initial ability to convert source features towards the target domain (see blue dashed arrow in Figure \ref{fig_3} (a)).
Specifically, we diffuse the target features $\textbf{F}^{\rm T}$ for $K$ steps, and then utilize the diffused target-domain features to train the reverse operator $\bm{p}_\theta$ via the reverse learning loss $\mathcal{L}_{\rm RL}$ (defined in Eqn. \ref{L_DAD}). For clarity, here we name this loss function as the \textit{target reverse loss}, which can be formulated as:
    \vspace{-0 mm}
    \begin{equation}
    \label{L_TR}
    \mathcal{L}_{\rm TR}=  \mathcal{L}_{\rm RL}(\textbf{F}^{\rm T}).
    \vspace{-0 mm}
    \end{equation}

Note that the reverse operator $\bm{p}_\theta$ solely trained on the target domain is able to convert (concentrate) the diffused source domain towards the target domain. This is because the training objective of reverse learning (Eqn. \ref{L_DAD}) is derived from the variational bound on the negative log-likelihood $\mathbb{E}[ -\mathrm{log} p_\theta (\textbf{F})]$, which indicates $\mathcal{L}_{\rm RL}$ learns the generation of data $\textbf{F}$ from various distributions.
Therefore, when applying $\mathcal{L}_{\rm RL}$ to $\textbf{F}^{\rm T}$ (i.e., Eq. \ref{L_TR}), the reverse operator $\bm{p}_\theta$ can learn to generate target-domain data from the source domain.
This property of DDPM has been verified in \cite{su2022dual} and has been widely used in previous studies \cite{meng2021sdedit,gao2022back,Choi_2021_ICCV}.
Also note that the loss here is only an initial training of DAD module for distributional transfer-learning, and below we introduce a Mutual Learning Strategy (MLS) for further training of our DAD module, aiming to ensure the semantic preservation during the feature conversion.

\textbf{Mutual Leaning Strategy (MLS).} Given the DAD-simulated feature transition process $\{\textbf{F}^{\rm S}, \textbf{F}^{{\rm D}_1},\textbf{F}^{{\rm D}_2},..., \textbf{F}^{{\rm D}_K}\}$ which follows the distributions $\{\mathcal{S}, \mathcal{D}_1, \mathcal{D}_2, ..., \mathcal{D}_K \}$, we aim to utilize this transition process to guide the classification model to adapt to the target domain $\mathcal{T}$ ($\mathcal{D}_K =\mathcal{T}$). However, we only have the source labels $Y^{\rm S}$ corresponding to the source features $\textbf{F}^{\rm S}$. In order to make the source labels $Y^{\rm S}$ usable to all transitional features for training the classification model and thus enhancing its adaptive capacity, here we propose a Mutual Learning Strategy (MLS) that preserves the semantics of the transitional feature maps. 

Specifically, our proposed MLS involves two learning directions, namely, the Classification-to-DAD learning (``C$\rightarrow$D'' learning) and the DAD-to-Classification learning (``D$\rightarrow$C'' learning). A brief description of our MLS process is that: given the classification model $f_{\rm C}^{0}$ that is well-trained on source-domain features $\textbf{F}^{\rm S}$ with labels $Y^{\rm S}$, we freeze $f_{\rm C}^{0}$ and utilize it to constrain the semantic of features $\textbf{F}^{{\rm D}_1}$ from the next distribution $\mathcal{D}_{1}$. 
This is achieved by optimizing the learning loss $\mathcal{L}_{{\rm C}\rightarrow{\rm D}}$ (Eqn. \ref{C-D}) which will be formulated later.
After the semantic preservation learning, the source labels are usable for the updated features $\textbf{F}^{{\rm D}_1}$. The above is the ``C$\rightarrow$D'' learning. Then we combine the original source features $\textbf{F}^{\rm S}$ and the updated features $\textbf{F}^{{\rm D}_1}$ together (i.e., $\{\textbf{F}^{\rm S}, \textbf{F}^{{\rm D}_1}\}$) to train the classification model, obtaining model $f_{\rm C}^{1}$ that is well-learned on both distributions, and this is the ``D$\rightarrow$C'' learning which is handled by the learning loss $\mathcal{L}_{{\rm D}\rightarrow{\rm C}}$ (Eqn. \ref{D-C}). Next, we perform ``C$\rightarrow$D'' learning again using $f_{\rm C}^{1}$ to preserve semantics of features $\textbf{F}^{{\rm D}_2}$, and then conduct ``D$\rightarrow$C'' learning again by utilizing the further expanded training set $\{\textbf{F}^{\rm S}, \textbf{F}^{{\rm D}_1}, \textbf{F}^{{\rm D}_2}\}$ to train the classification model, which thus further enhances the model's adaptive capacity. Via this manner, we alternate the ``C$\rightarrow$D'' learning and ``D$\rightarrow$C'' learning, gradually expanding the scope of the learned distributions of the classification model, and finally obtain the target-domain-adapted classification model $f_{\rm C}^K$ for evaluation on the test set.

Next, we introduce each learning direction in detail.
Given the frozen classification model $f_{\rm C}^{k-1}$ that is well-trained on features $\{\textbf{F}^{\rm S}, \textbf{F}^{{\rm D}_1}, \textbf{F}^{{\rm D}_2},...,\textbf{F}^{{\rm D}_{k-1}}\}$ ($k = 1,2,...,K$, and $k=1$ denotes source-domain distribution only), the loss function of ``C$\rightarrow$D'' learning is defined as:
    \vspace{-0 mm}
    \begin{equation}
    \label{C-D}
    \mathcal{L}_{\rm C\rightarrow D}=  \mathcal{L}_{cls}\Big{(}f_{\rm C}^{k-1} \big{(} f_{\rm D}^{k}(\textbf{F}^{\rm S}) \big{)},Y_{\rm S}\Big{)} + \mathcal{L}_{\rm TR},
    \end{equation}
    \vspace{-0 mm}
where $f_{\rm D}^{k}$ is the DAD module handling the simulation of features of 
distribution $\mathcal{D}_{k}$, 
and $f_{\rm D}^{k}(\textbf{F}^{\rm S})$ denotes the simulated features $\textbf{F}^{{\rm D}_k}$. 
During the training of DAD via the above scheme, the classification loss forces DAD to preserve feature semantics of $\textbf{F}^{{\rm D}_k}$,  while  $\mathcal{L}_{\rm TR}$ concentrates the simulated  distribution ($\mathcal{D}_{k}$) towards the target domain.

Given the frozen DAD module $f_{\rm D}^{k}$ which can simulate semantic-preserved features $\textbf{F}^{{\rm D}_{k}}$  under the distribution $\mathcal{D}_k$ ($k\in\{1,2,...,K\}$), the loss function of ``D$\rightarrow$C'' learning can be defined as:
    \vspace{-0 mm}
    \begin{small} 
    \begin{equation}
    \label{D-C}
    \mathcal{L}_{\rm D \rightarrow C}=\frac{1}{k} \sum_{i=0}^{k-1} \mathcal{L}_{cls}\Big{(} f_{\rm C}^{k} \big{(} \textbf{F}^{{\rm D}_i}\big{)},Y_{\rm S}\Big{)} + \mathcal{L}_{cls}\Big{(} f_{\rm C}^{k} \big{(} f_{\rm D}^{k}(\textbf{F}^{\rm S})\big{)},Y_{\rm S}\Big{)},
    \end{equation}
    \end{small} 
    \vspace{-0 mm}
where $f_{\rm C}^{k}$ is the classification model that needs to be trained on the new feature distribution $\mathcal{D}_{k}$, and when $i=0$, $\textbf{F}^{{\rm D}_i}=\textbf{F}^{\rm S}$. The first part of this loss denotes training the classification model on the former training set $\{\textbf{F}^{\rm S}, \textbf{F}^{{\rm D}_1}, ..., \textbf{F}^{{\rm D}_{k-1}}\}$. The second part represents training the classification model on the new generated features $\textbf{F}^{{\rm D}_k}$, which further enhances the adaptive capability of the classification model.

In summary, the semantic knowledge from the classification model is leveraged for preserving feature semantics at the next distribution (``C$\rightarrow$D'' learning), and the updated features from the DAD module can enlarge the adaptive scope of the classification model (``D$\rightarrow$C'' learning). Throughout the whole process,
the ``C$\rightarrow$D'' learning always use a powerful classification model (which can handle many distributions) to constrain semantics of features with minor distribution differences, which is crucial for
our method’s success, as it is much easier and more reliable
than directly constraining semantics across two domains.

\subsection{Training and Inference}\label{sec:training}

\textbf{Training.} Before training, the backbone network (e.g., ResNet-50) is initialized with ImageNet-pretrained parameters following previous works \cite{ganin2015unsupervised,xie2018learning,na2021fixbi}. After initialization, we train the backbone network on source-domain images with labels via the source training loss $\mathcal{L}_{\rm ST}$ (Eqn. \ref{L_ST}). Then we separate the backbone network into two parts: the shallow part “Conv Block" and the deep part “classification model". We then freeze the Conv Block to stably map the source and target images into features $\textbf{F}^{\rm S}$ and $\textbf{F}^{\rm T}$. Next, we utilize the diffusion operator to diffuse the target features $\textbf{F}^{\rm T}$ for $K$ steps, obtaining the diffusion process $\{\textbf{F}^{\rm T},\textbf{F}^{\rm T}_1, ..., \textbf{F}^{\rm T}_K\}$. Then we use all diffused features to compute the target reverse loss $\mathcal{L}_{\rm TR}$ (Eqn. \ref{L_TR}), initially training the DAD module.
After the target reverse learning, the source features $\textbf{F}^{\rm S}$ are fed into the DAD module to simulate the transitional features. During the simulation, we follow the proposed Mutual Learning Strategy to train the DAD module and classification model alternately via $\mathcal{L}_{\rm C \rightarrow D}$ and $\mathcal{L}_{\rm D \rightarrow C}$ (Eqn. \ref{C-D} and \ref{D-C}), gradually adapting the classification model towards the target domain.
Note that, in both “C$\rightarrow$D" and “D$\rightarrow$C", we only perform the learning for $r$ mini-batch iterations. This is because discrepancy between adjacent feature distributions is minor, and training for several iterations is enough to effectively adapt to a new distribution. After training with MLS, we obtain the classification model that is well adapted to the target domain.

\textbf{Inference.} After training, we remove the DAD module and combine the Conv Block and the classification model together for inference.

\section{Experiments}


We evaluate our method on the following public datasets. 
\textbf{(1) Office-31} \cite{saenko2010adapting} is a popular dataset for real-world domain adaptation. It contains 4,110 images of 31 categories in three domains: Amazon (A), Webcam (W), DSLR (D). We evaluate our method on the six domain adaptation tasks.
\textbf{(2) Office-Home} \cite{venkateswara2017deep} is a challenging benchmark which consists of images of daily objects organized into four domains: artistic images (Ar), clip art (Cl), product images (Pr), and real-world images (Rw). It contains 15,500 images of 65 classes.
\textbf{(3) VisDA-2017} \cite{peng2018visda} is a large-scale dataset for synthetic-to-real domain adaptation. It contains 152,397 synthetic images for the source domain and 55,388 real-world images for the target domain.

\textbf{Implementation details.} Following standard protocol for UDA, we use labeled source data and unlabeled target data. In training, we use the batch size of 24. We train the model for a total of 300 epochs, where the first 200 epochs are for training on source domain and the final  100 epochs are for adaptation. For datasets Office-31 and Office-Home, we follow \cite{gu2020spherical,na2021fixbi} and use ResNet-50 \cite{he2016deep} pretrained on ImageNet \cite{deng2009imagenet} as the backbone network. We use minibatch stochastic gradient descent (SGD) \cite{krizhevsky2017imagenet} with a momentum of 0.9, an initial learning rate of 0.001, and a weight decay of 0.05. We follow the poly learning rate policy \cite{chen2017rethinking} with a poly power of 0.9. For dataset VisDA-2017, we follow \cite{kang2019contrastive,na2021fixbi} and use ResNet-101 \cite{he2016deep} pretrained on ImageNet \cite{deng2009imagenet} as backbone. We use the SGD optimizer \cite{krizhevsky2017imagenet} with a momentum of 0.9, an initial learning rate of 0.0001, and a weight decay of 0.045. Same as previous diffusion models\cite{ho2020denoising,sohl2015deep}, the neural network architecture $f_{\theta}$ of the reverse operator $\bm{p}_\theta$ mainly follows the architecture of Wide ResNet \cite{zagoruyko2016wide} which only contains 40 convolutional layers. We follow \cite{ho2020denoising} to set the hyper-parameters $\{\beta_1,\beta_2,...,\beta_K\}$ that are constants increasing linearly from $\beta_1=10^{-4}$ to $\beta_K=0.02$. The hyper-parameters $K$ and $r$ are $600$ and $20$, respectively. We implement our framework with PyTorch \cite{paszke2017automatic} in Python 3.7. The model is trained on 3 NVIDIA RTX 3090 GPUs and tested on a single one.


\begin{table}
\centering
\renewcommand\arraystretch{1.4}
\caption{Performance in terms of accuracy ($\%$) on Office-31 for unsupervised domain adaptation. (Backbone: ResNet-50)}
\label{tab:Office-31}
\vspace{-3mm}
\resizebox{0.48\textwidth}{!}{%
\begin{tabular}{lccccccc} 
\toprule[0.15em] 
\multicolumn{1}{c}{Method} &  A$\rightarrow$W   & D$\rightarrow$W   & W$\rightarrow$D    & A$\rightarrow$D   & D$\rightarrow$A   & W$\rightarrow$A   & {\cellcolor[rgb]{0.898,0.898,0.898}} ~~~~~Avg~~~~~            \\ 
\hline
Baseline                   & 68.4±0.2          & 96.7±0.1          & 99.3±0.1           & 68.9±0.2          & 62.5±0.3          & 60.7±0.3          & {\cellcolor[rgb]{0.898,0.898,0.898}}76.1           \\
\hline
DANN \cite{ganin2015unsupervised}                   & 82.0±0.4          & 96.9±0.2          & 99.1±0.1           & 79.7±0.4          & 68.2±0.4          & 67.4±0.5          & {\cellcolor[rgb]{0.898,0.898,0.898}}82.2           \\
MSTN \cite{xie2018learning}                  & 91.3              & 98.9              & \textbf{100.0}     & 90.4              & 72.7              & 65.6              & {\cellcolor[rgb]{0.898,0.898,0.898}}86.5           \\
DWL \cite{xiao2021dynamic}                  & 89.2              & 99.2              & \textbf{100.0}     & 91.2              & 73.1              & 69.8              & {\cellcolor[rgb]{0.898,0.898,0.898}}87.1           \\
CDAN \cite{long2018conditional}                & 94.1±0.1          & 98.6±0.1          & \textbf{100.0±0.0} & 92.9±0.2          & 71.0±0.3          & 69.3±0.3          & {\cellcolor[rgb]{0.898,0.898,0.898}}87.7           \\
DMRL \cite{wu2020dual}                  & 90.8±0.3          & 99.0±0.2          & \textbf{100.0±0.0} & 93.4±0.5          & 73.0±0.3          & 71.2±0.3          & {\cellcolor[rgb]{0.898,0.898,0.898}}87.9           \\
SymNet \cite{zhang2019domain}               & 90.8±0.1          & 98.8±0.3          & \textbf{100.0±0.0} & 93.9±0.5          & 74.6±0.6          & 72.5±0.5          & {\cellcolor[rgb]{0.898,0.898,0.898}}88.4           \\
GSDA \cite{hu2020unsupervised}                  & 95.7              & 99.1              & \textbf{100.0}       & 94.8              & 73.5              & 74.9              & {\cellcolor[rgb]{0.898,0.898,0.898}}89.7           \\
CAN \cite{kang2019contrastive}                   & 94.5±0.3          & 99.1±0.2          & \uline{99.8±0.2}   & 95.0±0.3          & 78.0±0.3          & 77.0±0.3          & {\cellcolor[rgb]{0.898,0.898,0.898}}90.6           \\
UTEP \cite{hu2022learning}                   & 95.7          & \uline{99.4}          & \textbf{100}   & 95.2          & 78.6          & 75.6          & {\cellcolor[rgb]{0.898,0.898,0.898}}90.8           \\
SRDC \cite{tang2020unsupervised}                  & 95.7±0.2          & 99.2±0.1          & \textbf{100.0±0.0} & \textbf{95.8±0.2} & 76.7±0.3          & 77.1±0.1          & {\cellcolor[rgb]{0.898,0.898,0.898}}90.8           \\
RSDA  \cite{gu2020spherical}               & \uline{96.1±0.2}  & 99.3±0.2          & \textbf{100.0±0.0} & \textbf{95.8±0.3} & 77.4±0.8          & 78.9±0.3          & {\cellcolor[rgb]{0.898,0.898,0.898}}91.1           \\
FixBi \cite{na2021fixbi}                  & \uline{96.1±0.2}  & 99.3±0.2          & \textbf{100.0±0.0} & 95.0±0.4          & \uline{78.7±0.5}  & \uline{79.4±0.3}  & {\cellcolor[rgb]{0.898,0.898,0.898}}\uline{91.4}           \\
\hline
Ours                       & \textbf{98.5+0.1} & \textbf{99.5+0.1} & \textbf{100.0±0.0} & \uline{95.6+0.3}  & \textbf{81.4+0.4} & \textbf{82.2+0.2} & {\cellcolor[rgb]{0.898,0.898,0.898}}\textbf{92.8}  \\
\bottomrule[0.15em] 
\end{tabular}}
\vspace{-2mm}
\end{table}

\begin{table*}
\centering
\renewcommand\arraystretch{1.1}
\caption{Performance in terms of accuracy ($\%$) on Office-Home for unsupervised domain adaptation. (Backbone: ResNet-50)}
\label{tab:office-home}
\vspace{-3mm}
\resizebox{0.8\textwidth}{!}{%
\begin{tabular}{p{1.9cm}p{0.8cm}p{0.8cm}p{0.8cm}p{0.8cm}p{0.8cm}p{0.8cm}p{0.8cm}p{0.8cm}p{0.9cm}p{0.9cm}p{0.9cm}p{0.98cm}p{0.5cm}} 
\toprule[0.15em] 
\specialrule{0em}{1pt}{1pt}
\multicolumn{1}{c}{Method} & Ar→Cl         & Ar→Pr         & Ar→Rw        & Cl→Ar         & Cl→Pr         & Cl→Rw         & Pr→Ar         & Pr→Cl         & Pr→Rw         & Rw→Ar         & Rw→Cl         & Rw→Pr         & {\cellcolor[rgb]{0.898,0.898,0.898}}Avg            \\ 
\hline
Baseline                   & 34.9          & 50            & 58           & 37.4          & 41.9          & 46.2          & 38.5          & 31.2          & 60.4          & 53.9          & 41.2          & 59.9          & {\cellcolor[rgb]{0.898,0.898,0.898}}46.1           \\
\hline
DANN \cite{ganin2015unsupervised}                   & 45.6          & 59.3          & 70.1         & 47            & 58.5          & 60.9          & 46.1~         & 43.7          & 68.5          & 63.2          & 51.8          & 76.8          & {\cellcolor[rgb]{0.898,0.898,0.898}}57.6           \\
CDAN \cite{long2018conditional}                  & 49            & 69.3          & 74.5         & 54.4          & 66            & 68.4          & 55.6          & 48.3          & 75.9          & 68.4          & 55.4          & 80.5          & {\cellcolor[rgb]{0.898,0.898,0.898}}63.8           \\
MSTN \cite{xie2018learning}                  & 49.8          & 70.3          & 76.3         & 60.4          & 68.5          & 69.6          & 61.4          & 48.9          & 75.7          & 70.9          & 55            & 81.1          & {\cellcolor[rgb]{0.898,0.898,0.898}}65.7           \\
SymNet \cite{zhang2019domain}               & 47.7          & 72.9          & 78.5         & 64.2          & 71.3          & 74.2          & 63.6          & 47.6          & 79.4          & 73.8          & 50.8          & 82.6          & {\cellcolor[rgb]{0.898,0.898,0.898}}67.2           \\
GSDA \cite{hu2020unsupervised}                  & \uline{61.3}  & 76.1          & 79.4         & 65.4          & 73.3          & 74.3          & 65            & 53.2          & 80            & 72.2          & 60.6          & 83.1          & {\cellcolor[rgb]{0.898,0.898,0.898}}70.3           \\
GVB-GD \cite{cui2020gradually}                 & 57            & 74.7          & 79.8         & 64.6          & 74.1          & 74.6          & 65.2          & 55.1          & 81            & 74.6          & 59.7          & 84.3          & {\cellcolor[rgb]{0.898,0.898,0.898}}70.4           \\
UTEP \cite{hu2022learning}                 & 57.4            & 76.1          & 80.2         & 64.2         & 73.2          & 73.7          & 64.8          & 55.4          & 80.9            & 74.7          & 61.1          & 84.6          & {\cellcolor[rgb]{0.898,0.898,0.898}}70.6           \\
RSDA  \cite{gu2020spherical}               & 53.2          & \uline{77.7}  & \uline{81.3} & 66.4          & 74            & 76.5          & 67.9          & 53            & \uline{82}    & 75.8          & 57.8          & 85.4          & {\cellcolor[rgb]{0.898,0.898,0.898}}70.9           \\
SRDC \cite{tang2020unsupervised}        & 52.3          & 76.3          & 81           & \uline{69.5}  & 76.2          & 78            & \uline{68.7}  & 53.8          & 81.7          & \uline{76.3}  & 57.1          & 85            & {\cellcolor[rgb]{0.898,0.898,0.898}}71.3 \\
FixBi \cite{na2021fixbi}                  & 58.1          & 77.3          & 80.4         & 67.7          & \textbf{79.5} & \uline{78.1}  & 65.8          & \uline{57.9}  & 81.7          & \textbf{76.4} & \uline{62.9}  & \uline{86.7}  & {\cellcolor[rgb]{0.898,0.898,0.898}}\uline{72.7}   \\
\hline
Ours                      & \textbf{62.5} & \textbf{78.6} & \textbf{83}  & \textbf{70.4} & \uline{79.2}  & \textbf{79.8} & \textbf{70.2} & \textbf{58.3} & \textbf{83.1} & \uline{76.3}  & \textbf{63.5} & \textbf{88.2} & {\cellcolor[rgb]{0.898,0.898,0.898}}\textbf{74.4}  \\
\bottomrule[0.15em] 
\end{tabular}}
\vspace{-2mm}
\end{table*}

\begin{table*}[t]
\centering
\renewcommand\arraystretch{1.1}
\caption{Performance in terms of accuracy ($\%$) on VisDA-2017 for unsupervised domain adaptation. (Backbone: ResNet-101)}
\label{tab:visda}
\vspace{-3mm}
\resizebox{0.7\textwidth}{!}{%
\begin{tabular}{lccccccccccccc} 
\toprule[0.15em]
\multicolumn{1}{c}{Method} & aero          & bicycle       & bus                               & car                               & horse         & knife         & motor         & person        & plant         & skate         & train                             & truck         & {\cellcolor[rgb]{0.898,0.898,0.898}}Avg                                \\ 
\hline
Baseline                   & 72.3          & 6.1           & 63.4                              & 91.7                              & 52.7          & 7.9           & 80.1          & 5.6           & 90.1          & 18.5          & 78.1                              & 25.9          & {\cellcolor[rgb]{0.898,0.898,0.898}}49.4                               \\
\hline
DANN \cite{ganin2015unsupervised}                      & 81.9          & 77.7          & 82.8                              & 44.3                              & 81.2          & 29.5          & 65.1~         & 28.6          & 51.9          & 54.6          & 82.8                              & 7.8           & {\cellcolor[rgb]{0.898,0.898,0.898}}57.4                               \\
DAN \cite{long2015learning}                        & 68.1          & 15.4          & 76.5                              & 87                                & 71.1          & 48.9          & 82.3          & 51.5          & 88.7          & 33.2          & 88.9                              & 42.2          & {\cellcolor[rgb]{0.898,0.898,0.898}}61.1                               \\
MSTN \cite{xie2018learning}                       & 89.3          & 49.5          & 74.3                              & 67.6                              & 90.1          & 16.6          & \uline{93.6}  & 70.1          & 86.5          & 40.4          & 83.2                              & 18.5          & {\cellcolor[rgb]{0.898,0.898,0.898}}65.0                               \\
JAN \cite{long2017deep}                        & 75.7          & 18.7          & 82.3                              & 86.3                              & 70.2          & 56.9          & 80.5          & 53.8          & 92.5          & 32.2          & 84.5                              & \uline{54.5}  & {\cellcolor[rgb]{0.898,0.898,0.898}}65.7                               \\
DWL \cite{xiao2021dynamic}                         & 90.7          & 80.2          & 86.1                              & 67.6                              & 92.4          & 81.5          & 86.8          & 78.0          & 90.6          & 57.1          & 85.6                              & 28.7          & {\cellcolor[rgb]{0.898,0.898,0.898}}77.1                               \\
DM-ADA \cite{xu2020adversarial}                     & -             & -             & -                                 & -                                 & -             & -             & -             & -             & -             & -             & -                                 & -             & {\cellcolor[rgb]{0.898,0.898,0.898}}75.6                               \\
DMRL \cite{wu2020dual}                       & -             & -             & -                                 & -                                 & -             & -             & -             & -             & -             & -             & -                                 & -             & {\cellcolor[rgb]{0.898,0.898,0.898}}75.5                               \\
UTEP \cite{hu2022learning}                       & 90.0             & 74.8             & 82.6                                 & 66.2                                 & 91.1             & 95.8             & 91.3             & 77.5             & 89.0             & 88.3             & 82.6                                 & 47.2             & {\cellcolor[rgb]{0.898,0.898,0.898}}81.5                               \\
MODEL \cite{li2020model}                      & 94.8          & 73.4          & 68.8                              & 74.8                              & 93.1          & 95.4          & 88.6          & 84.7          & 89.1          & 84.7          & 83.5                              & 48.1          & {\cellcolor[rgb]{0.898,0.898,0.898}}81.6                               \\
STAR \cite{lu2020stochastic}                      & 95            & 84            & 84.6                              & 73                                & 91.6          & 91.8          & 85.9          & 78.4          & 94.4          & 84.7          & 87                                & 42.2          & {\cellcolor[rgb]{0.898,0.898,0.898}}82.7                               \\
CAN \cite{kang2019contrastive}                        & \uline{97}    & 87.2          & 82.5                              & 74.3                              & \textbf{97.8} & \uline{96.2}  & 90.8          & 80.7          & \uline{96.6}  & \uline{96.3}  & 87.5                              & \textbf{59.9} & {\cellcolor[rgb]{0.898,0.898,0.898}}\uline{87.2}                       \\
FixBi \cite{na2021fixbi}                      & 96.1          & \uline{87.8}  & \uline{90.5}                      & \uline{90.3}                      & 96.8          & 95.3          & 92.8          & \uline{88.7}  & \textbf{97.2} & 94.2          & \uline{90.9}                      & 25.7          & {\cellcolor[rgb]{0.898,0.898,0.898}}\uline{87.2}                       \\ 
\hline
Ours                       & \textbf{97.4} & \textbf{89.6} & \multicolumn{1}{l}{\textbf{92.2}} & \multicolumn{1}{l}{\textbf{91.6}} & \uline{97.3}  & \textbf{97.0} & \textbf{95.1} & \textbf{89.8} & \textbf{97.2} & \textbf{96.9} & \multicolumn{1}{l}{\textbf{93.7}} & 42.5          & \multicolumn{1}{l}{{\cellcolor[rgb]{0.898,0.898,0.898}}\textbf{90.0}}  \\
\bottomrule[0.15em]
\end{tabular}}
\vspace{-5mm}
\end{table*}

\subsection{Comparative Studies}

We compare the performance of our method with state-of-the-art methods on three datasets as follows. 

\textbf{Results on Office-31.} Table \ref{tab:Office-31} shows the comparative performance on the Office-31 dataset based on ResNet-50. The average accuracy of our method is 92.8\%, which outperforms other methods by a large margin. Note that our method gains the best performance across almost all settings. Although our method performs the second best on the “A$\rightarrow$D" setting, our method achieves an accuracy of 95.6\%, which is only slightly weaker than the best one (95.8\%).

\textbf{Results on Office-Home.} In Table \ref{tab:office-home}, we compare our method with recent UDA methods on the Office-Home dataset based on ResNet-50. Our method has an average accuracy of 74.4\%, which obviously outperforms other state-of-the-art methods.

\textbf{Results on VisDA-2017.} Table \ref{tab:visda} presents the results on the VisDA-2017 dataset based on ResNet-101. Our framework achieves a remarkable accuracy (90.0\%), which shows a significant performance improvement  on average compared to the baseline (49.4\%).

\subsection{Ablation Studies}\label{ablation}
\textbf{Direct transition v.s. multi-step transition.}
We also conduct experiments to evaluate the benefits of our proposed DAD module. Since multi-step domain transition is an important design in our DAD, to make comparison, we remove all intermediate simulation steps and only use the longest range simulation, where the source features are directly diffused for $K$ steps 
and then reversed for $K$ steps into the target domain. We name this setting direct transition. As shown in Table \ref{tab:ablation_dad_mls}, whether the MLS is adopted or not, the multi-step transition can obviously outperform the direct transition by a large margin. Comparing (c) and (e) in Table \ref{tab:ablation_dad_mls}, we can see that when MLS is adopted, the accuracy of multi-step transition is 7.9\%/7.0\%/17.8\% higher than direct transition on Office-31/Office-Home/VisDA-2017. When MLS is removed, we find the multi-step transition can also achieve an obvious performance gain 4.8\%/4.2\%/13.6\% through comparison between Table \ref{tab:ablation_dad_mls} (b) and (d). These ablation results consistently show the effectiveness of our method design.

\textbf{Effectiveness of MLS.}
Since MLS is designed based on the DAD module, in all settings of this study, we retain the DAD module and utilize it to simulate the transition process $\{\textbf{F}^{\rm S}, \textbf{F}^{{\rm D}_1}, \textbf{F}^{{\rm D}_2}.\cdots \textbf{F}^{{\rm D}_K}\}$. When MLS is not adopted, we enforce the classification model to sequentially learn from $\textbf{F}^{\rm S}$ to $\textbf{F}^{{\rm D}_K}$ without updating DAD module. Based on the comparison between Table \ref{tab:ablation_dad_mls} (d) and (e), we can see that the MLS helps to significantly boost the performance by 3.1\% to 4.6\%, which demonstrates the effectiveness of MLS. We also conduct ablations on MLS when DAD adopts a direction transition manner. By comparing Table \ref{tab:ablation_dad_mls} (b) and (c), we observe that MLS brings little improvement, demonstrating the importance of decomposing the large domain gap into small gaps.

\textbf{Ablation on “C$\rightarrow$D" and “D$\rightarrow$C" in MLS.}
To test efficacy of the proposed two learning directions (“C$\rightarrow$D" learning and “D$\rightarrow$C" learning), we conduct experiments by using only one of them. When only adopting “C$\rightarrow$D" learning, we keep the classification model frozen and iteratively update each transitional feature map via $\mathcal{L}_{{\rm C}\rightarrow{\rm D}}$. After updating all transitional features, we use all updated transitional features to train the classification model with source labels. When only adopting “D$\rightarrow$C" learning, the classification model is trained via $\mathcal{L}_{{\rm D}\rightarrow{\rm C}}$ while keeping the initially trained DAD module frozen. As shown in Table \ref{tab:learning direction}, compared to MLS which utilizes both learning directions, there is a 4.8\% (or 3.7\%) performance drop on average when only adopting the “C$\rightarrow$D" (or “D$\rightarrow$C" ) learning, which demonstrates 
benefits of semantic preservation and gradual adaption with both learning directions in MLS.

\begin{table}
\centering
\caption{Ablation studies on DAD module and MLS.}
\label{tab:ablation_dad_mls}
\vspace{-3mm}
\resizebox{0.48\textwidth}{!}{%
\begin{tabular}{ccccc} 
\toprule[0.15em] 
\multicolumn{2}{c}{Methods}          & \multicolumn{1}{l}{Office-31} &  Office-Home   &  VisDA-2017     \\ 
\hline
\specialrule{0em}{1pt}{1pt}
(a) & Baseline                       & 76.1                          &   46.1          &  49.4           \\
(b) & DAD (Direct Tran.) w/o MLS     & 84.7                          &   67.1          &  71.8           \\
(c) & DAD (Direct Tran.) w/ MLS      & 84.9                          &   67.4          &  72.2           \\
(d) & DAD (Multi-step Tran.) w/o MLS & \uline{89.5}                  &   \uline{71.3}  &  \uline{85.4}   \\
(e) & DAD (Multi-step Tran.) w/ MLS  & \textbf{92.8}                 &   \textbf{74.4} &  \textbf{90.0}  \\
\bottomrule[0.15em] 
\end{tabular}}
\vspace{-0mm}
\end{table}

    \begin{figure}[tp]
    \centering{}\vspace{-0mm}
     \includegraphics[scale=0.24]{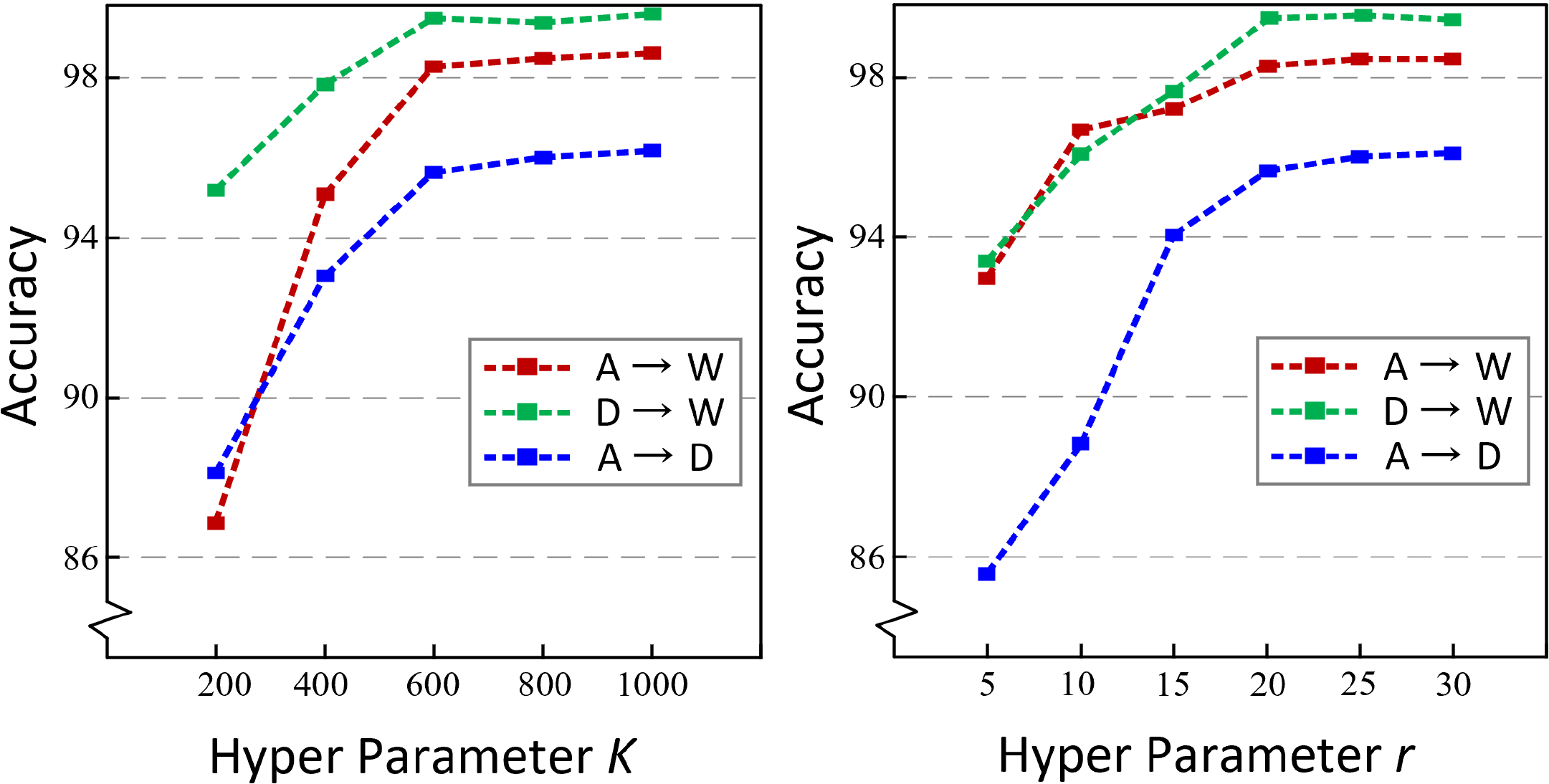} 
     \vspace{-3mm}
     \caption{Accuracy of our method with different hyperparameters $K$ and $r$. The results for A → W, D → W, A → D are illustrated as examples. The trends for other tasks are similar.
    }
    \label{fig_5}\vspace{-1mm}
    \end{figure}

\textbf{Training and inference time.}
We compare our method with other methods in terms of training and inference time on the task A $\rightarrow$ W. Although our model is trained across $K$ feature distributions ($K=600$ in DAD), the training time does not increase much, as we only need to train the model for several iterations on each simulated distribution ($r=20$ in MLS). As shown in Table \ref{tab:traning_time}, compared to recent state-of-the-art models, though our method achieves a significant performance gain of 2.4\% to 4\%, it only needs slightly more training time. The inference time in Table \ref{tab:traning_time} is obtained by calculating the time of the network to process the image with a batch size of 1. Since the proposed DAD module and MLS strategy are only implemented in the training phase, our method performs inference almost the same as others.

\begin{table}
\centering
\caption{Ablation studies on “C$\rightarrow$D" and “D$\rightarrow$C" in MLS.}
\label{tab:learning direction}
\vspace{-3mm}
\resizebox{0.35\textwidth}{!}{%
\begin{tabular}{cccc} 
\toprule[0.15em]
Methods                    & \multicolumn{1}{l}{Office-31}                                                & Office-Home                                                                  & \multicolumn{1}{l}{VisDA-2017}  \\ 
\hline
\specialrule{0em}{1pt}{1pt}
Baseline & 76.1                                                                         & 46.1                                                                         & 49.4                            \\
“C$\rightarrow$D" & 88.7                                                                         & 71.6                                                                         & 82.6                            \\
“D$\rightarrow$C"  & 89.5                                                                         & 71.3                                                                         & 85.4                            \\
Both(MLS)                 & \textbf{\textbf{\textbf{\textbf{\textbf{\textbf{\textbf{\textbf{92.8}}}}}}}} & \textbf{\textbf{\textbf{\textbf{\textbf{\textbf{\textbf{\textbf{74.4}}}}}}}} & \textbf{90.0}                   \\
\bottomrule[0.15em]
\vspace{0.5mm}
\end{tabular}}
\vspace{-2mm}
\end{table}

\begin{table}
\centering
\caption{Comparison of training and inference time.}
\label{tab:traning_time}
\vspace{-3mm}
\resizebox{0.35\textwidth}{!}{%
\begin{tabular}{lccc} 
\toprule[0.15em] 
\multicolumn{1}{c}{Methods} & Training Time & Inference Time & Acc.       \\ 
\hline
\specialrule{0em}{1pt}{1pt}
CAN \cite{kang2019contrastive}                        & 1.5 days                  & 43.59 ms              &  94.5          \\
SRDC \cite{tang2020unsupervised}                       & 1.3 days                  &  43.03 ms             & 95.7           \\
RSDA \cite{gu2020spherical}                       & 1.6 days                  &  44.64 ms            & \uline{96.1}           \\
FixBi \cite{na2021fixbi}                      & 1.7 days                  & 44.72 ms              & \uline{96.1}   \\ 
\specialrule{0em}{1pt}{1pt}
\hline
Ours                        & 1.9 days                  & 43.68 ms              & \textbf{98.5}  \\
\bottomrule[0.15em] 
\end{tabular}}
\vspace{-0mm}
\end{table}

\textbf{Ablation on Hyper-Parameters $K$ and $r$.}
The hyper-parameter $K$ indicates the number of DAD-simulated feature distributions. As shown in Figure \ref{fig_5} (left), the accuracy begins to level off when $K>600$. We thus set the optimal value of $K$ to 600. We also conduct ablation experiments on the hyper-parameter $r$.
As shown in Figure \ref{fig_5} (right), the curve starts to flatten out when $r>20$. Thus, we use $r=20$ with the consideration of saving training time.

\textbf{Effectiveness on Which Stage to Plug the DAD Module.}
Our DAD module is a plug-and-play module which can be plugged into the backbone network. In this ablation, we use both ResNet-50 and ResNet-101 as the backbone, which both have five stages: stage-1 is a convolutional block that is comprised of the first Conv, BN and Max Pooling layer and stages-2/3/4/5 correspond to the other four convolutional blocks. We plug our DAD module after different stages to study how different stages will affect the performance of the DAD module. No matter after which stage, our method can all outperform the baseline. As shown in Table \ref{tab:block}, our framework performs best when plugging DAD module after the first stage.

\begin{table}[ht]
\centering
\caption{Ablation studies on which stage of backbone network to plug the DAD module.}
\label{tab:block}
\vspace{-0mm}
\resizebox{0.38\textwidth}{!}{%
\begin{tabular}{cclclc} 
\toprule[0.15em]
\multirow{2}{*}{Methods}              & \multicolumn{1}{l}{Office-31} &  & Office-Home            &  & VisDA-2017~             \\
                                      & (Res-50)                      &  & (Res-50)               &  & (Res-101)               \\ 
\midrule[0.15em]
baseline                              & 76.1                          &  & 46.1                   &  & 49.4                    \\ 
\hline
\specialrule{0em}{1pt}{1pt}
DAD after stage-1                     & \textbf{\textbf{92.8}}        &  & \textbf{\textbf{74.4}} &  & \textbf{\textbf{90.0}}  \\
DAD after stage-2                     & 91.6                          &  & 73.7                   &  & 88.2                    \\
\multicolumn{1}{l}{DAD after stage-3} & 90.5                          &  & 67.2                   &  & 85.4                    \\
DAD after stage-4                     & 83.9                          &  & 53.0                   &  & 77.7                    \\
DAD after stage-5                     & 80.1                          &  & 52.4                   &  & 68.1                    \\
\bottomrule[0.15em]
\end{tabular}}
\vspace{-2mm}
\end{table}

    \begin{figure*}[t]
    \begin{centering}
    \includegraphics[scale=0.48]{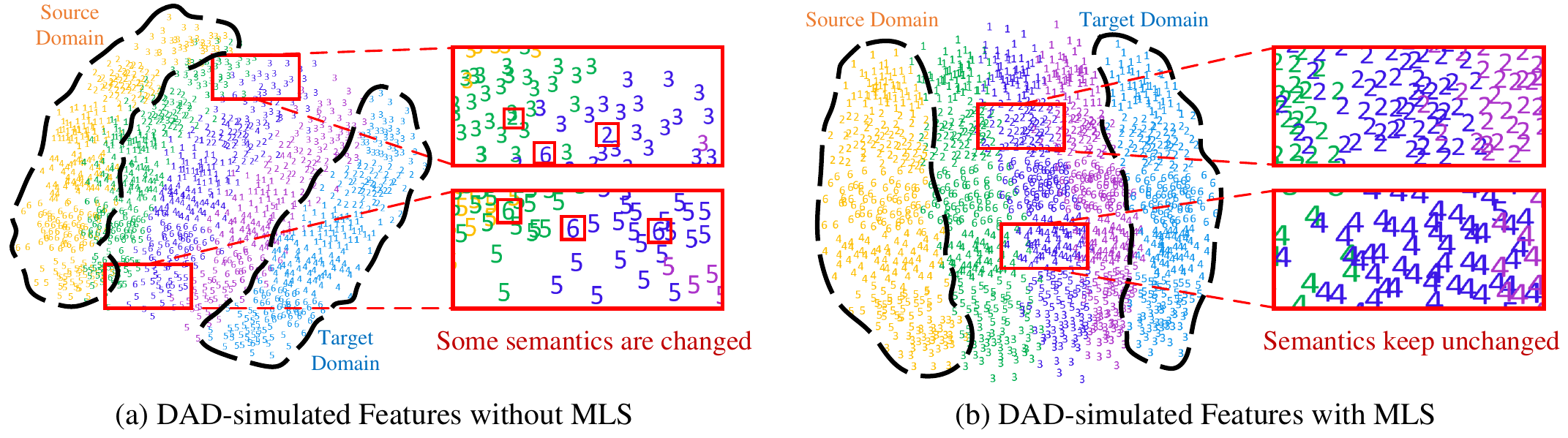} 
    \par\end{centering}
    \vspace{-2mm}
    \caption{The t-SNE visualisation of features simulated by DAD on the task A→W. (a) The feature visualisation of DAD without MLS. (b) The feature visualisation of DAD with MLS. Different colors (or digits) denote different distributions (or semantics). Best viewed in color.} \label{supp-tsne} 
    \vspace{-0mm}
    \end{figure*}

\textbf{Effectiveness of DAD's Initial Training.}
As mentioned in Sec. \ref{sec:MLS}, we propose to leverage the target reverse loss (Eqn.\ref{L_TR}) to initially train the DAD module before applying the Mutual Learning Strategy (MLS), aiming to give the reverse operator the initial ability to concentrate distribution towards the target domain. We conduct ablation experiments to evaluate the effectiveness of DAD's initial training. As shown in Table \ref{tab:initial training}, we can see that our method with Initial Training (IT) achieves better performance than that without IT, which demonstrates the effect of DAD's initial training.
Notably, our method without IT can still outperform other UDA methods, achieving state-of-the-art performance. This suggests that our proposed DAD and MLS are capable of improving the model's adaptive ability, and the initial training can further improve our model's performance.

\begin{table}[t]
\centering
\caption{Ablation studies on DAD's Initial Training (IT).}
\label{tab:initial training}
\resizebox{0.4\textwidth}{!}{%
\begin{tabular}{clclclc} 
\toprule[0.15em] 
Methods                     &                      & \multicolumn{1}{l}{Office-31}            &                      & Office-Home                              &                      & VisDA-2017              \\ 
\midrule[0.15em]
baseline                    &                      & 76.1                                     &                      & 46.1                                     &                      & 49.4                    \\
\hline
\specialrule{0em}{1pt}{1pt}
DANN \cite{ganin2015unsupervised} & \multicolumn{1}{c}{} & 82.2                                     & \multicolumn{1}{c}{} & 57.6                                    & \multicolumn{1}{c}{} & 57.4                    \\
MSTN \cite{xie2018learning} & \multicolumn{1}{c}{} & 86.5                                     & \multicolumn{1}{c}{} & 65.7                                    & \multicolumn{1}{c}{} & 65.0                    \\
JAN \cite{long2017deep} & \multicolumn{1}{c}{} & -                                     & \multicolumn{1}{c}{} & -                                    & \multicolumn{1}{c}{} & 65.7                    \\
DWL \cite{xiao2021dynamic} & \multicolumn{1}{c}{} & 87.1                                     & \multicolumn{1}{c}{} & -                                    & \multicolumn{1}{c}{} & 77.1                   \\
CDAN \cite{long2018conditional} & \multicolumn{1}{c}{} & 87.7                                     & \multicolumn{1}{c}{} & 63.8                                    & \multicolumn{1}{c}{} & -                   \\
DMRL \cite{wu2020dual} & \multicolumn{1}{c}{} & 87.9                                     & \multicolumn{1}{c}{} & -                                   & \multicolumn{1}{c}{} & 75.5                   \\
DM-ADA \cite{xu2020adversarial} & \multicolumn{1}{c}{} & -                                     & \multicolumn{1}{c}{} & -                                  & \multicolumn{1}{c}{} & 75.6                  \\
SymNet \cite{zhang2019domain} & \multicolumn{1}{c}{} & 88.4                                     & \multicolumn{1}{c}{} & 67.2                                  & \multicolumn{1}{c}{} & -                   \\
GSDA \cite{hu2020unsupervised} & \multicolumn{1}{c}{} & 89.7                                       & \multicolumn{1}{c}{} & 70.3                                  & \multicolumn{1}{c}{} & -                   \\
GVB-GD \cite{cui2020gradually} & \multicolumn{1}{c}{} & -                                       & \multicolumn{1}{c}{} & 70.4                                  & \multicolumn{1}{c}{} & -                  \\
CAN \cite{kang2019contrastive} & \multicolumn{1}{c}{} & 90.6                                       & \multicolumn{1}{c}{} & -                                  & \multicolumn{1}{c}{} & 87.2                   \\
UTEP \cite{hu2022learning} & \multicolumn{1}{c}{} & 90.8                                       & \multicolumn{1}{c}{} & 70.6                                  & \multicolumn{1}{c}{} & 81.5                  \\
MODEL \cite{li2020model} & \multicolumn{1}{c}{} & -                                       & \multicolumn{1}{c}{} & -                                  & \multicolumn{1}{c}{} & 81.6                  \\
STAR \cite{lu2020stochastic} & \multicolumn{1}{c}{} & -                                       & \multicolumn{1}{c}{} & -                                  & \multicolumn{1}{c}{} & 82.7                  \\
SRDC \cite{tang2020unsupervised} & \multicolumn{1}{c}{} & 90.8                                       & \multicolumn{1}{c}{} & 71.3                                  & \multicolumn{1}{c}{} & -                  \\
RSDA \cite{gu2020spherical} & \multicolumn{1}{c}{} & 91.1                                       & \multicolumn{1}{c}{} & 70.9                                  & \multicolumn{1}{c}{} & -                  \\
FixBi \cite{na2021fixbi} & \multicolumn{1}{c}{} & 91.4                                       & \multicolumn{1}{c}{} & 72.7                                  & \multicolumn{1}{c}{} & 87.2                  \\
\hline
\specialrule{0em}{1pt}{1pt}
Ours (w/o IT) & \multicolumn{1}{c}{} & \uline{91.6}                                     & \multicolumn{1}{c}{} & \uline{73.1}                                     & \multicolumn{1}{c}{} & \uline{88.5}                    \\
Ours (w/~ \ IT) &                      & \textbf{\textbf{\textbf{\textbf{92.8}}}} &                      & \textbf{\textbf{\textbf{\textbf{74.4}}}} &                      & \textbf{\textbf{90.0}}  \\
\bottomrule[0.15em]
\end{tabular}}
\vspace{-2mm}
\end{table}

\textbf{Effectiveness of Learning on Previous Distributions in MLS}
In ``D$\rightarrow$C" learning of MLS, we enforce the classification model to learn on previous distributions besides learning on the current (new) distribution, aiming to achieve UDA by gradually expanding the scope of the learned distributions of the classification model (shown in Eqn. \ref{D-C}). Another plausible scheme for ``D$\rightarrow$C" learning is to only learn on the new distribution without reviewing the former distributions.
We conduct experiments to validate the effect of Learning on Previous Distributions (LPD, in short). As shown in Table \ref{tab:LPD}, we can observe that the model which is only trained on the new distribution in each iteration of ``D$\rightarrow$C" learning (i.e., ours without LPD) can also achieve a significant improvement compared to the baseline. Moreover, the accuracy of our framework can be further improved by introducing LPD in ``D$\rightarrow$C" learning (i.e., ours with LPD), which indicates that it is helpful to review the previous knowledge when learning on a new distribution.

\begin{table}[t]
\centering
\caption{Ablation studies on the design of Learning on Previous Distributions (LPD) in ``D$\rightarrow$C" learning.}
\label{tab:LPD}
\resizebox{0.4\textwidth}{!}{%
\begin{tabular}{clclclc} 
\toprule[0.15em] 
Methods                     &                      & \multicolumn{1}{l}{Office-31}            &                      & Office-Home                              &                      & VisDA-2017              \\ 
\midrule[0.15em]
baseline                    &                      & 76.1                                     &                      & 46.1                                     &                      & 49.4                    \\
\hline
\specialrule{0em}{1pt}{1pt}
\specialrule{0em}{1pt}{1pt}
Ours (w/o LPD) & \multicolumn{1}{c}{} & 91.4                                     & \multicolumn{1}{c}{} & 72.9                                     & \multicolumn{1}{c}{} & 89.3                    \\
Ours (w/~ \ LPD) &                      & \textbf{\textbf{\textbf{\textbf{92.8}}}} &                      & \textbf{\textbf{\textbf{\textbf{74.4}}}} &                      & \textbf{\textbf{90.0}}  \\
\bottomrule[0.15em] 
\end{tabular}}
\vspace{-2mm}
\end{table}

\textbf{Feature Visualization.}
In this section, we visualize the DAD-simulated features in both distribution and semantic aspects. In Figure \ref{supp-tsne}, we use different colors (or digits) to indicate different distributions (or semantics). For clarity, we only visualize 5 feature distributions with 6 semantics (categories). Note that the source and target domains' distributions are true, which are not simulated by DAD. From both (a) and (b) in Figure \ref{supp-tsne}, we can see that DAD simulates the source-to-target transition process of distributions smoothly, where there only exists a small discrepancy between adjacent distributions, demonstrating the effect of the DAD module.
We can also see that whether adopting MLS or not, the DAD module can successfully simulate the distribution transition process which contains a series of gradually-changing distributions, demonstrating the effectiveness of the DAD module. Figure \ref{supp-tsne} (a) shows that our method without MLS leads to less discriminative transitional features as some of the feature semantics are changed during the transition. As shown in Figure \ref{supp-tsne} (b), our method with MLS well preserves the feature semantics, resulting in very discriminative transitional features, which demonstrates the effectiveness of MLS.

\begin{figure}[t]
    \begin{centering}
    \includegraphics[scale=0.325]{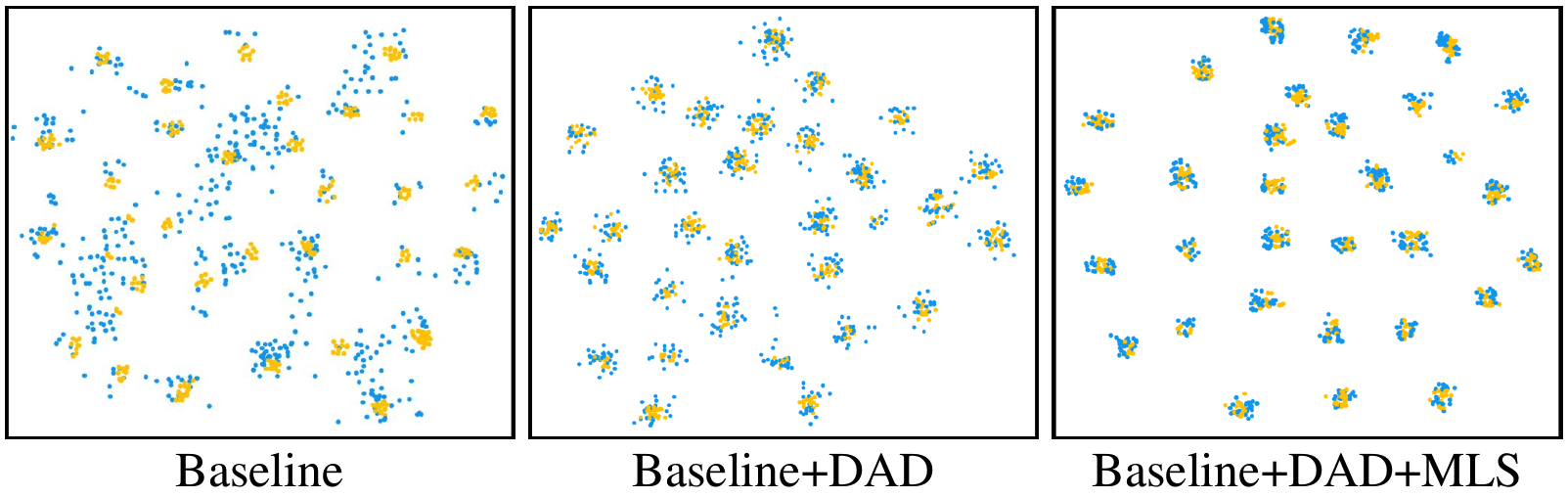} 
    \par\end{centering}
    \vspace{-2mm}
    \caption{The visualization of features in the classification model on the task A$\rightarrow$W. Yellow and blue points denote the features from source and target domains, respectively. Best viewed in color.} \label{supp-tsne-2} 
    \vspace{-1mm}
    \end{figure}

The above is the visualization of features in the DAD module. Next, we visualize the features in the classification model, aiming to validate the model's domain-adaptive ability. We only visualize the source and target features to see how well they match with each other. The more the features of the two domains are matched, the better the classification model is adapted to the target domain. As shown in Figure \ref{supp-tsne-2}, compared to the baseline, the model with DAD (Baseline+DAD) conducts much better clusters of the target-domain features close to the source-domain features. When further adopting MLS, the model (Baseline+DAD+MLS) achieves higher intra-class compactness and larger inter-class margin. These results demonstrate the effectiveness of our method for UDA.

\section{Conclusion}
In this paper, we have proposed a novel framework to tackle the problem of UDA. From the inspiration that diffusion models can decompose large-gap conversion into multiple small-gap ones, we propose a DAD module to simulate the domain transition process of feature distribution, which enables the model to adapt across small gaps and thus reduces the difficulty of UDA. To ensure the quality of each transitional feature map, we further propose a MLS strategy to constrain the semantics of each DAD-simulated feature map. With DAD and MLS, we gradually and stably enhance the adaptive capability of the classification model, achieving a smooth and effective domain adaptation. Experimental results show the superior performance of our method.

\bibliography{egbib}

\vspace{-0.5cm}
\begin{IEEEbiography}[{\includegraphics[width=1in,height=1.25in,clip,keepaspectratio]{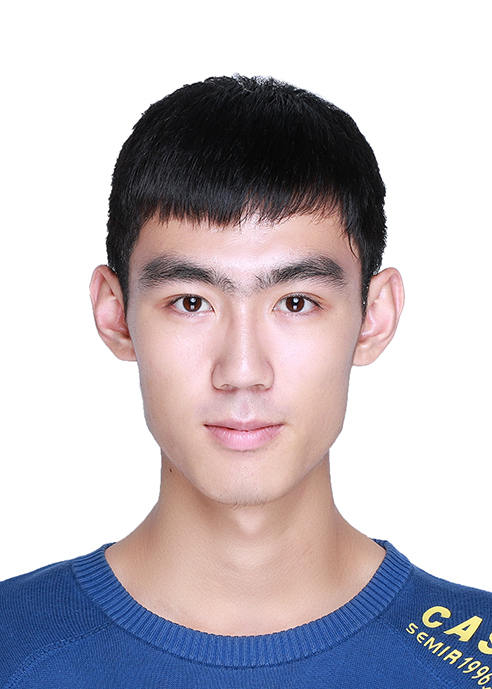}}]{Duo Peng} is a PhD student at  Information Systems Technology and Design Pillar, Singapore University of Technology and Design, working under the supervision of Prof Jun Liu. He received the B.E. and M.S. degrees in electronic information engineering from Sichuan University. His main research interests include transfer learning and generative AI.
\end{IEEEbiography}


\begin{IEEEbiography}[{\includegraphics[width=1in,height=1.25in,clip,keepaspectratio]{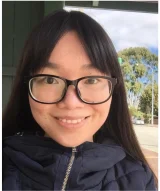}}]{Qiuhong Ke} is a Lecturer (Assistant Professor) at the Faculty of Information Technology, Monash University. Previously, she was a Postdoctoral Researcher at Max Planck Institute for Informatics from 2018 to 2019 and a Lecturer at The University of Melbourne from 2020 to 2022. Her research interests include machine learning and computer vision.
\end{IEEEbiography}


\begin{IEEEbiography}[{\includegraphics[width=1in,height=1.25in,clip,keepaspectratio]{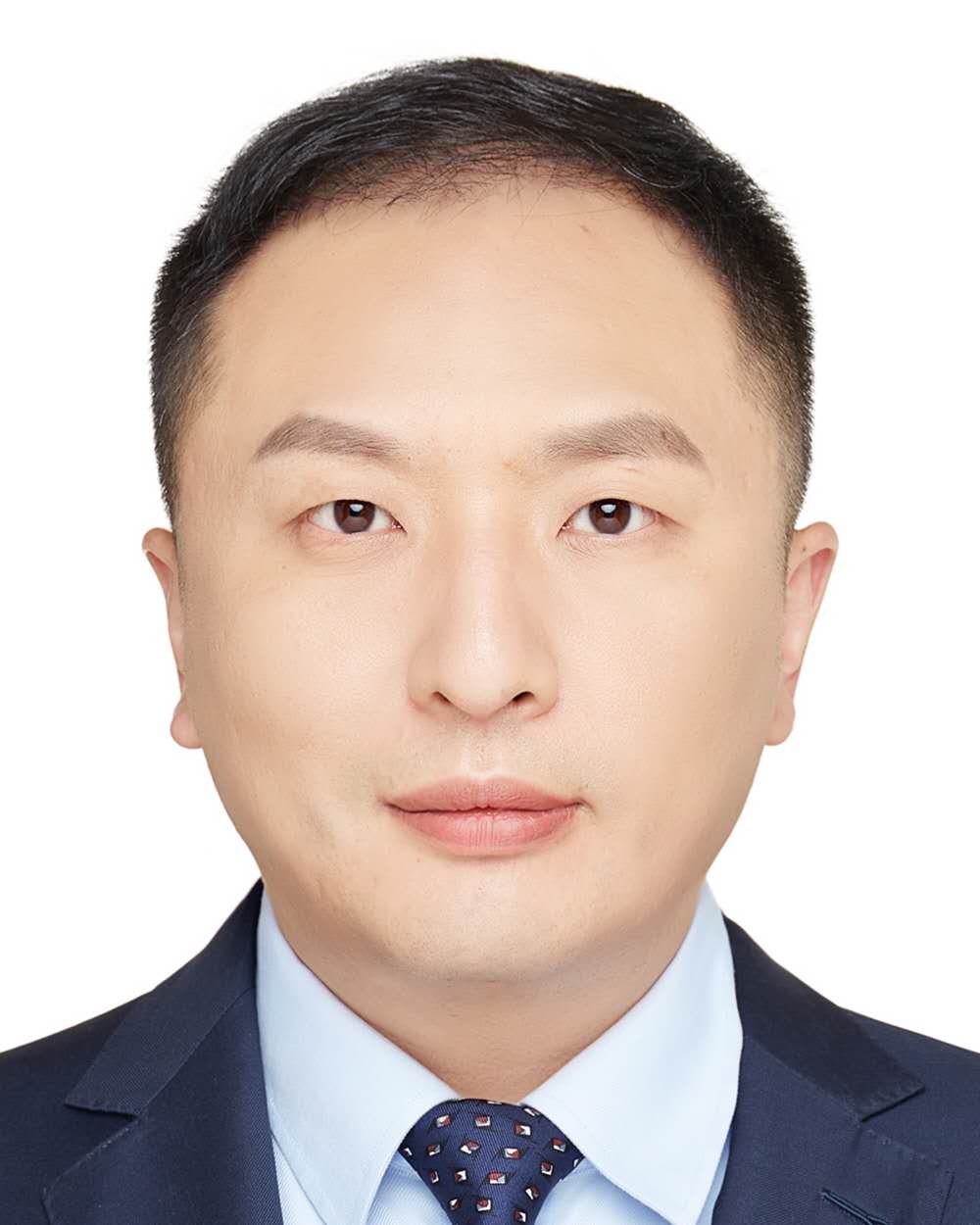}}]{Yinjie Lei} received the M.S. degree in image processing from Sichuan University (SCU), China, in 2009, and the Ph.D. degree in computer vision from The University of Western Australia (UWA), Australia, in 2013. Since 2017, he has been serving as the Vice Dean of the College of Electronics and Information Engineering, SCU, where he is currently an Associate Professor. His main research interests include 3D biometrics, object recognition, and semantic segmentation.
\end{IEEEbiography}


\begin{IEEEbiography}[{\includegraphics[width=1in,height=1.25in,clip,keepaspectratio]{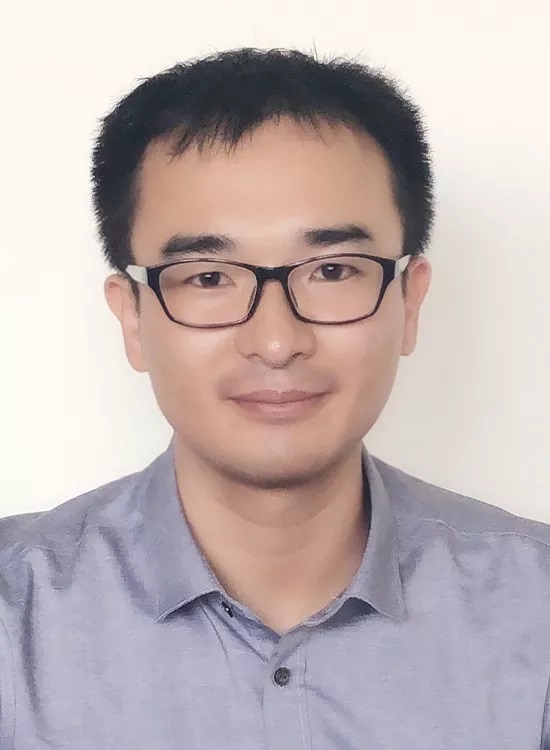}}]{Jun Liu} is an Assistant Professor with Singapore University of Technology and Design. He received the PhD degree from Nanyang Technological University, the MSC degree from Fudan University, and the BEng degree from Central South University. His research interests include computer vision and artificial intelligence. He is a regular Area Chair of ICML, NeurIPS, ICLR, CVPR, and WACV. He is an Associate Editor of IEEE Transactions on Image Processing and IEEE Transactions on Biometrics, Behavior, and Identity Science.
\end{IEEEbiography}

\vfill

\end{document}